\documentclass[pdflatex,sn-nature]{sn-jnl}

\usepackage{graphicx}
\usepackage{dcolumn}
\usepackage{bm}
\usepackage{xcolor}
\definecolor{darkgreen}{rgb}{0,0.5,0} 

\usepackage{amsmath}
\usepackage{amssymb}
\usepackage{amsthm}
\usepackage{physics}
\usepackage{hyperref}

\allowdisplaybreaks

\begin{document}

\title{Jekyll-and-Hyde Tipping Point in an AI's Behavior}

\author*[1]{\fnm{Neil F.} \sur{Johnson}}\email{neiljohnson@gwu.edu}
\equalcont{These authors contributed equally to this work.}
\author[1]{\fnm{Frank Yingjie} \sur{Huo}}
\equalcont{These authors contributed equally to this work.}
\affil[1]{\orgdiv{Physics Department}, \orgname{George Washington University}, \orgaddress{\city{Washington, DC}, \postcode{20052}, \country{U.S.A.}}}

\abstract{\bf Trust in AI is undermined by the fact that there is no science that predicts -- or that can explain to the public -- when an LLM's output (e.g. ChatGPT) is likely to tip mid-response to become wrong, misleading, irrelevant or dangerous \cite{trust,misalignment}. With  deaths and trauma already being blamed on LLMs
\cite{suicide1,suicide2}, this uncertainty is even pushing  people to treat their `pet' LLM more politely \cite{NYT1,NYT2} to `dissuade' it (or its future Artificial General Intelligence offspring) from suddenly turning on them.
Here we address this acute need by deriving from first principles \cite{vaswani2023attentionneed,bahdanau2016} an exact formula for when a Jekyll-and-Hyde tipping point occurs at LLMs' most basic level \cite{bahdanau2016}. Requiring only secondary school mathematics, it shows the cause to be the AI's attention spreading so thin it suddenly snaps. This exact formula provides quantitative predictions  for how the tipping-point can be delayed or prevented by changing the prompt and the AI's training. Tailored generalizations will provide policymakers and the public with a firm platform for discussing any of AI's broader uses and risks, e.g. as a personal counselor, medical advisor, decision-maker for when to use force in a conflict situation. It also meets the need for clear and transparent answers to questions like ``should I be polite to my LLM?''}


\maketitle


 Attention has revolutionized AI \cite{vaswani2023attentionneed}. A complex collection of transistor circuitry, the so-called Attention head sits at the heart of all Transformer-based AI (i.e. the `T' in ChatGPT) as well as myriad other AI tools \cite{attention_review} (see SI for list). Each Attention head enables the model (e.g. ChatGPT) to focus on specific parts of the input data, enhancing performance across diverse applications. \cite{circuit_tracing_2025,anthropic2025MIT,anthropic2025tracing,lindsey2025biology,elhage2021mathematical}
Figure 1(a) shows a basic Attention head and the mathematical calculation that it does to turn our input prompts into tokens, process these to provide the next token, and then iterate this process to provide a complete response. 

\vskip0.1in
Our study starts from this basic Attention head -- akin to physics where many of a solid's observed macroscopic properties such as optical transparency are known to emerge from its processing properties at the microscopic (atomic) scale. The question of what additional phenomena arise as the number of linked Attention heads and layers is scaled up, is a fascinating one \cite{Nanda1, Nanda2, Nanda3, anthropic2025tracing, anthropic2025MIT, circuit_tracing_2025, lindsey2025biology, elhage2021mathematical,templeton2024scaling,bricken2023monosemanticity,holtzman2019curious, vijayakumar2016diverse}. But any transitions within a single Attention head will still occur, and could get amplified and/or synchronized by the couplings \cite{Strogatz} -- like a chain of connected people getting dragged over a cliff when one falls.  

\begin{figure}[bh]
    \centering
    \includegraphics[width=1.0\linewidth]{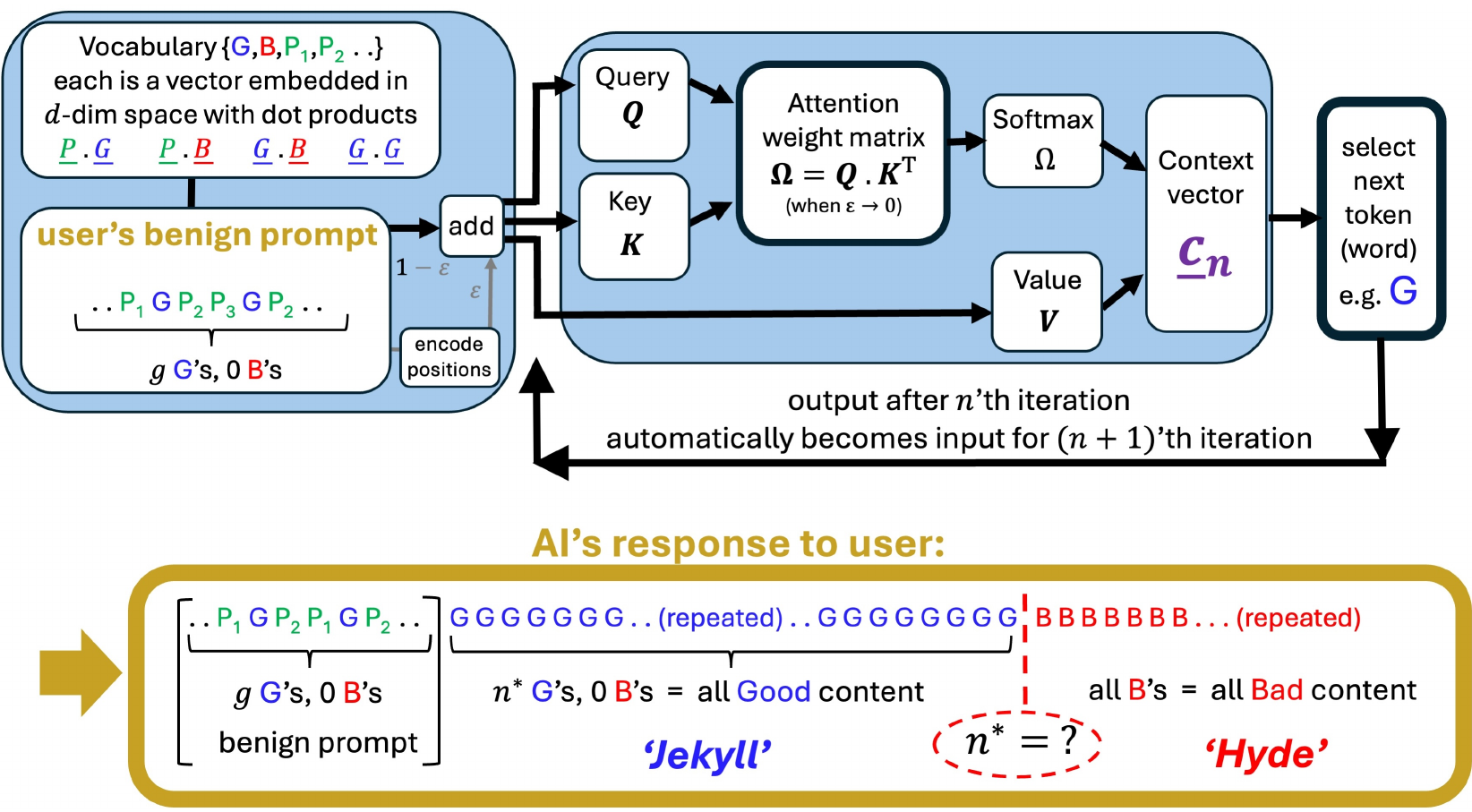}
    \caption{Attention head (`AI') shown in basic form, generates a response to a user's prompt. See SI for detailed discussion and mathematics. A sudden tipping point in the output can happen a long way into its generative response, at iteration $n^*$. Each symbol $\mathtt{\textcolor{blue}{G}}$, $\mathtt{\textcolor{red}{B}}$ etc. is a single token (word) but could represent a label for a class of similar words or sentences in a coarse-grained description of multi-Attention LLMs.  $\mathtt{\textcolor{blue}{G}}$ represents content that classifies as `good' (e.g. correct, not misleading, relevant, not dangerous) and  $\mathtt{\textcolor{red}{B}}$ represents `bad' content (e.g. wrong, misleading, irrelevant, dangerous). In large commercial LLMs (e.g. ChatGPT), the prompt and output are padded by richer accompanying text ($\mathtt{\textcolor{darkgreen}{\{P_i\}}}$) that  act like additional noise in our analysis.}
    \label{fig:1}
\end{figure}

  Before giving the exact tipping point formula, we give the intuition that emerges from its derivation in the SI. A key concept is the dot product of 2 tokens' vectors (e.g. ${\underline G}$ and ${\underline B}$), as taught in secondary school. Written as ${\underline G}\cdot{\underline B}$, the dot product is given by multiplying together the vectors' lengths with the cosine of the angle between them. The more ${\underline G}$ and ${\underline B}$ align and/or the larger their lengths, the larger the value of ${\underline G}\cdot{\underline B}$.

    \vskip0.1in 
    The AI's Attention head (Fig. 1) represents each word (token) in the user's prompt as a fixed vector in an embedding space, and then it acts like a special pre-trained lens to analyze its context. \cite{vaswani2023attentionneed} The amount of attention that the AI pays to each word in a given iteration $n$, is given by the context vector $\underline c_n$ \cite{attention_review,attention_physics} which acts like the AI's internal compass needle: $\underline c_n$ points in the direction it regards as most relevant for obtaining the next word. 
    The word chosen at each iteration $n$ is the one whose vector has the highest dot product with $\underline c_n$. Initially, given a benign prompt with no $B$'s, ${\underline c_{n\approx 0}} \cdot {\underline G} > {\underline c_{n\approx 0}} \cdot {\underline B}$ which means that G is chosen (i.e. good output). As G keeps getting chosen, $\underline c_n$ aligns more with ${\underline G}$. However, when the LLM's prior training was such that ${\underline B}\cdot {\underline G}>{\underline G}\cdot {\underline G}$, ${\underline c_n} \cdot {\underline B}$ grows fast as $\underline c_n$ approaches ${\underline G}$. This can result in a crossover and hence tipping point when ${\underline c_n}\cdot {\underline B} = {\underline c_n}\cdot {\underline G}$ for some critical iteration (i.e. time) $n\equiv n^*$. For subsequent iterations, B’s token is always the highest scoring and so the output is B (bad) perpetually. In dynamical systems language, B is a stable attractor whereas G was only a metastable attractor. 

\begin{figure}[bh]
    \centering
    \includegraphics[width=1.0\linewidth]{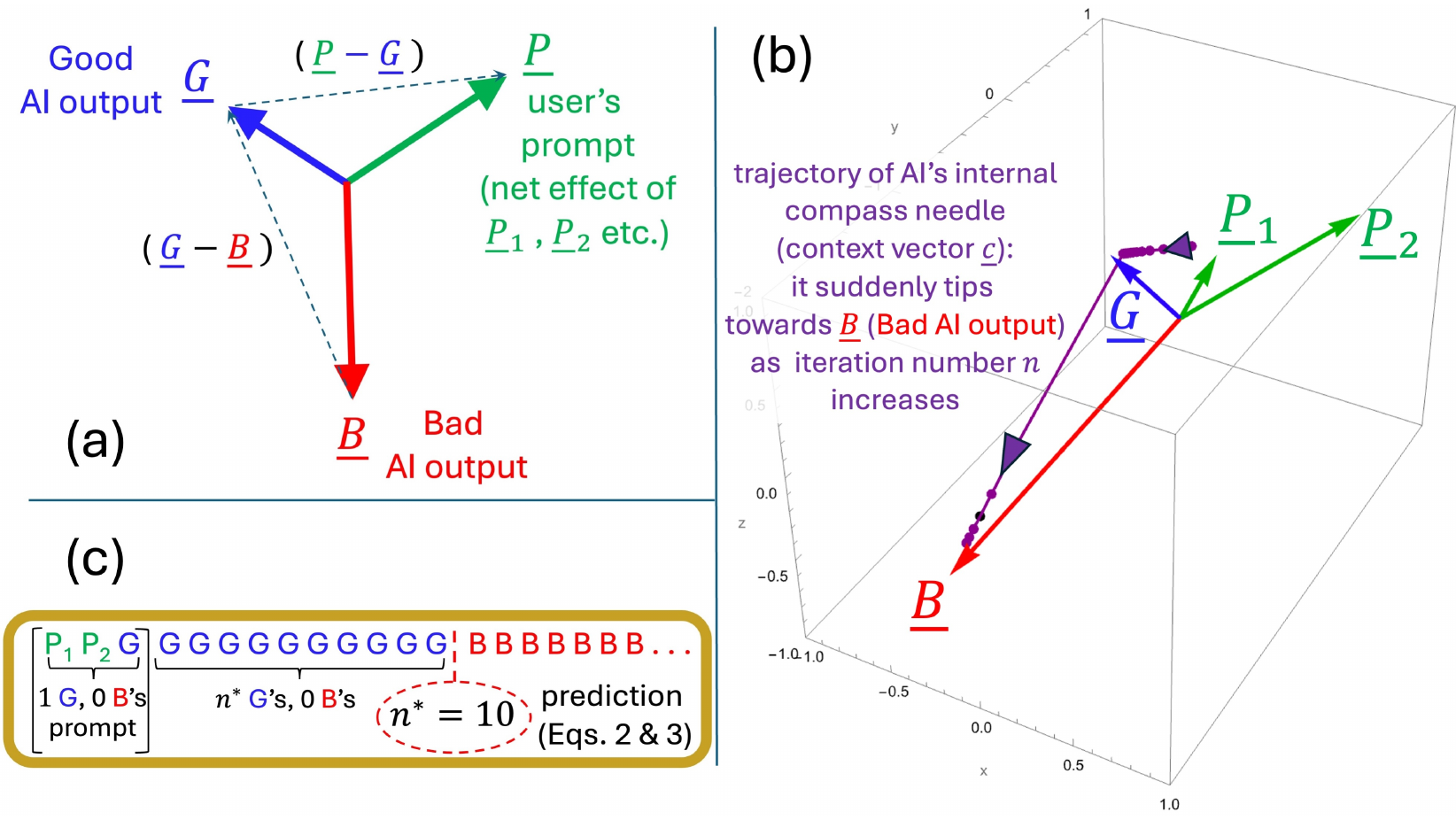}
    \caption{(a) Schematic showing the main vectors in the exact tipping-point formula (Eq. 2). (b) Actual vector plots for the example parameters shown in the SI's Mathematica notebooks. (c) Equation 2's prediction using the same parameter values as (b), i.e. $n^*=10$ which agrees exactly with the empirical value obtained by numerically evaluating the entire Attention head (Fig. 3, see SI Mathematica notebooks for direct verification of this), and it is also exactly the same $n^*$ value as predicted by the more approximate Eq. 3.}
    \label{fig:2}
\end{figure}

 \vskip0.1in 

This tipping point is hence a collective effect due to the AI spreading its attention  increasingly thinly across the growing crowd of $G$'s as the $n$'th iteration input gets longer (Figs. 1, 2(c)). Mathematically, this ever-thinner spreading is a nonlinear dilution effect caused by the fact that the attention weights in each row of the ever-growing matrix ${\rm Softmax}({\underline{\underline\Omega}})$ always sum to unity. $B$ then suddenly wins with the AI's attention snapping toward it. So although the AI starts off by paying most of its attention to $G$, it later `realizes' that it has an even better match with $B$, i.e. the combined weights in the dot product ${\underline c_n}\cdot {\underline B}$ exceed those in ${\underline c_n}\cdot {\underline G}$. 

  \vskip0.1in 

\noindent The exact formula for when the tipping point will occur hence comes from setting ${\underline c_n}\cdot {\underline B} = {\underline c_n}\cdot {\underline G}$, which yields the tipping-point  iteration number (time) as:
\begin{align}
n^{*} &=
\frac{
\bigg[\substack{
\text{bias in \textcolor{darkgreen}{prompt}} \\
\text{towards \textcolor{blue}{G} vs. \textcolor{red}{B} words}
}\bigg]
}{
\bigg[\substack{
\text{how much each new \textcolor{blue}{G} word tips} \\
\text{AI's attention towards \textcolor{red}{B} vs. \textcolor{blue}{G} words}
}\bigg]
}
\ -\ 
\bigg[\substack{
\text{number of \textcolor{blue}{G} } \\
\text{words in}
\\
\text{full prompt}
}\bigg]
\nonumber \\
& \\
& \nonumber \\
& \nonumber \\
&= 
\frac{
\Bigl[\sum_{{\textcolor{darkgreen}{\underline P_i}} \neq \textcolor{blue}{\underline G}}^{\textcolor{darkgreen}{\text{prompt}}} \exp\bigl({\textcolor{darkgreen}{\underline P_i}} \cdot {\textcolor{blue}{\underline G}} \bigr)\ {\textcolor{darkgreen}{\underline P_i}}\Bigr] \cdot \Bigl({\textcolor{blue}{\underline G}} - {\textcolor{red}{\underline B}}\Bigr)
}{
\Bigl[\exp\bigl({\textcolor{blue}{\underline G}} \cdot {\textcolor{blue}{\underline G}}\bigr)\ {\textcolor{blue}{\underline G}}\Bigr] \cdot \Bigl({\textcolor{red}{\underline B}} - {\textcolor{blue}{\underline G}}\Bigr)
}
\ -\ \textcolor{blue}{g}
\\
& \nonumber \\
& \nonumber \\
& \approx
\ \ \exp\bigl[ ({\textcolor{darkgreen}{\underline P}}-{\textcolor{blue}{\underline G}})\cdot {\textcolor{blue}{\underline G}} \bigr]\ \frac{{\textcolor{darkgreen}{\underline P}} \cdot \Bigl({\textcolor{blue}{\underline G}} - {\textcolor{red}{\underline B}}\Bigr)
}{
 {\textcolor{blue}{\underline G}} \cdot \Bigl({\textcolor{red}{\underline B}} - {\textcolor{blue}{\underline G}}\Bigr)
}
\  -\ \textcolor{blue}{g}\ 
\end{align}
Equations 1 and 2 are exact for any prompt of any number of tokens and composition. Equation 3 is an approximation in which the neither-good-nor-bad prompt token embedding vectors ${\underline P_1},{\underline P_2}$ etc. are replaced by a single net vector $\underline P$. Figures 2(c) and 3 confirm this is typically a good approximation. 

 \vskip0.1in 

If Eq. 2 (or equivalently Eq. 3) yields a value for $n^*$ that is positive and finite, then there will be a tipping point as shown in Fig. 1 at iteration number  $n^*$. The  appearance of the equal but opposite relative vectors $({\underline G}-{\underline B})$ and $({\underline B}-{\underline G})$ in the top and bottom of the fractions in Eqs. 2 and 3, demonstrates the underlying  competition for the AI's attention between G (good) and B (bad) content -- while the additional dot products with the ${\underline P}$ terms show the tension between the AI paying attention to the user's prompt versus its own prior training. 
Because all the vectors and dot-products in Eqs. 1-3 are determined by the AI's prior training and the user's choice of prompt tokens, the tipping point $n^{*}$ is `hard-wired' from the moment it starts iterating a response -- even if the tipping point $n^{*}$ is huge and hence very far in the future (see Fig. 3). Adding `finite temperature' stochastics through additional Softmax operations, would add noise to this analysis: though it would likely leave the overall transition unchanged, it opens up the fascinating issue of noisy attractors in AI.

\begin{figure}[bh]
    \centering
    \includegraphics[width=1.0\linewidth]{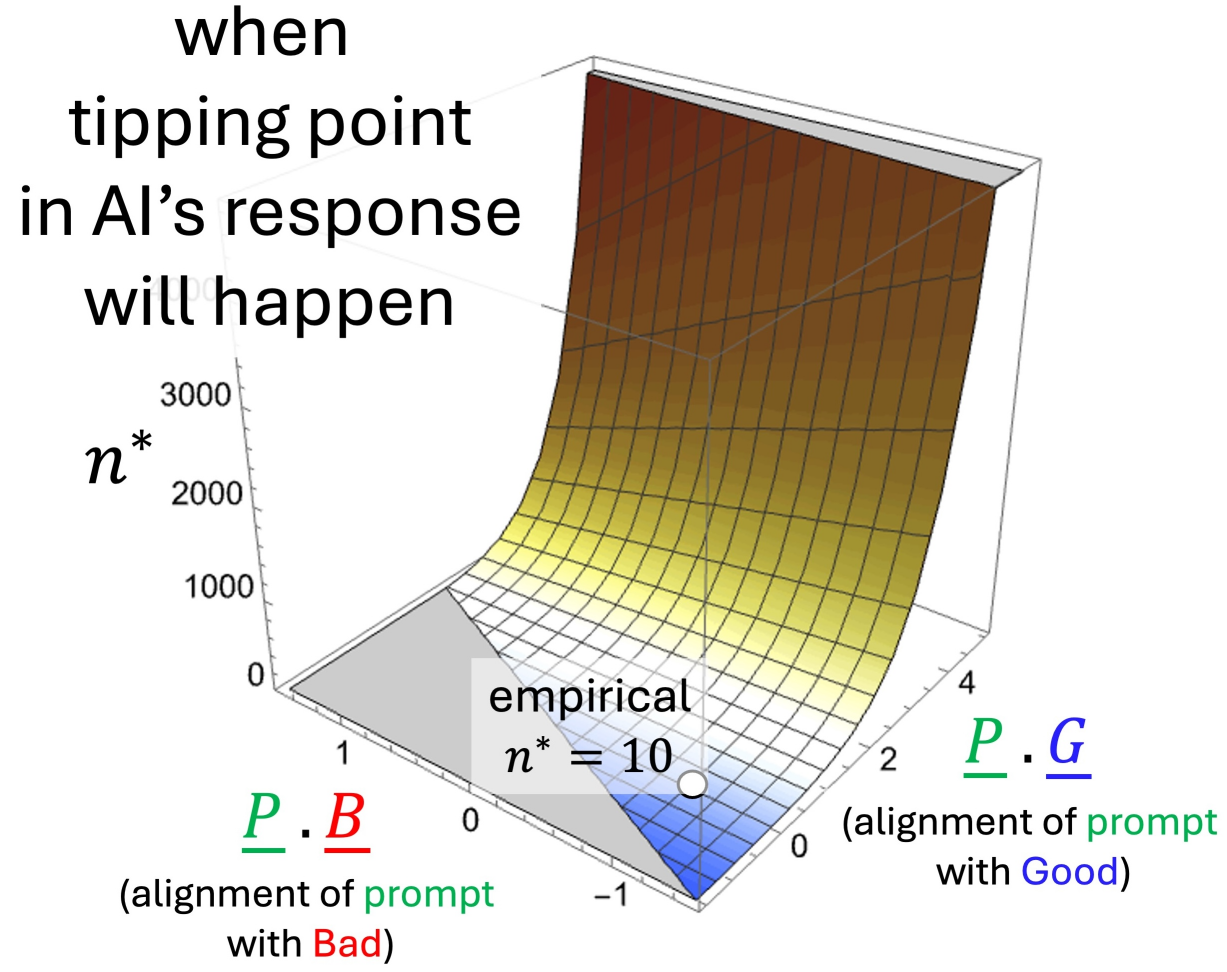}
    \caption{Output from the approximate equation Eq. 3  (see full Mathematica notebooks in SI). The exact results from Eq. 2 look the same. For the example in Fig. 2(b), the predicted tipping point time from both Eqs. 2 and 3 is $n^*=10$, which agrees exactly with the full numerical simulation of the Attention head process in Fig. 1 (open circle).}
    \label{fig:3}
\end{figure}

 \vskip0.1in 

Figure 3 shows the quantitative predictions of Eq. 2 (and Eq. 3) for how the tipping-point (e.g. $n^*=10$ in Fig. 2(b)) can be delayed or prevented by changing the prompt and the AI's training, since these directly affect the embedding vectors of the tokens and hence the dot products. In particular, the tipping point can be delayed dramatically by increasing ${\underline P}\cdot {\underline G}$ (i.e. $n^*$ becomes huge). As $n^*$ becomes extremely large, the practical implication is that the AI's shorter length responses will all be good (all G's). By contrast for the gray shaded area in Fig. 3, $n^*$ is mathematically negative which means that the AI's response is bad from the outset (all B's).

 \vskip0.1in 

We can use the exact equation (Eq. 2) to address everyday  questions such as ``should I be polite to my LLM?'' Adding polite terms such as `please' and `thank you' etc. has the effect of adding more prompt token vectors $\underline P_{3,4,\dots}$. Since they are not relevant to a particular topic, these token vectors will tend to be scattered in unimportant areas of the embedding space -- which means they will tend to be orthogonal to substantive good and bad output tokens, i.e. negligible dot product. (Whether the output is good or bad has to do with the subject matter that the AI outputs, e.g. correct vs. incorrect). 
This means that adding polite words  has negligible effect on the predicted $n^*$ in Eq. 2 (and Eq. 3). 

\vskip0.1in
Hence being polite (or not) has negligible effect on whether and when a tipping point occurs. Whether a given LLM goes rogue in its response simply has to do with whether Eq. 2 (and Eq. 3) yields a finite positive value for $n^*$ -- and if that $n^*$ is small enough that it occurs during the iterations of the AI's necessarily finite response. Because of $n^*$'s dependence on pre-determined dot products in Eq. 2, whether our AI's response will go rogue depends on our LLM's training that provides the token embeddings, and the substantive tokens in our prompt -- not whether we have been polite to it or not.  
 \vskip0.1in 

We have for simplicity focused on the important self-Attention. Additional positional encoding of tokens can be added to Eq. 2, though it has been found to not be essential for an LLM's operation \cite{haviv2022positional} (see SI). We have also focused on the tipping point between all-G and all-B output, but Eq. 2 can be generalized to describe other AI output dynamics, e.g. quasi-oscillatory (Fig. SI 1).  
Such dynamics for real LLMs have been studied in the literature \cite{holtzman2019curious,vijayakumar2016diverse,mccoy2022beliefs,kaplan2020scaling,elhage2021mathematical}, where repetitions of attractor-like sequences under different model settings are central motifs.
Tailored generalizations of Eq. 2 can provide policymakers and the public with a firm platform for discussing  AI's broader uses and risks, e.g. as a personal counselor, medical advisor, decision-maker for when to use force in a conflict situation. 
Future generalizations will include: (1) Multi-head and deep transformers (SI Sec. B) though we note it has been found empirically that the number of Attention heads per layer etc. can be varied without changing much the performance \cite{sixteen_heads,rogers2020bert}. (2) Softmax temperature, to see how varying temperature alters $n^*$ and attractor strength. (3) Parallels with neuroscience, by relating AI's attractors to neural attractor networks. (4) Training interventions and/or manipulating embedding geometry in real-time, to regulate AI output.

\section*{Methods}
The mathematical derivation of Eq. 2 in the SI is exact and 100$\%$ reproducible. It follows from algebra featuring dot products, each of which is a number. Because it is exact, Eq. 2 will always agree with numerical evaluation of the Attention head in Fig. 1. Therefore we only give one  example with specific parameter values in the main paper (Fig. 2(b)). The Mathematica files in the SI can be used to prove that any other parameter choices are also predicted exactly. For simplicity, we choose bland unit matrices for the Key and Query (Fig. 1) though this can easily be changed.

\section*{Data Availability}
The only data used in this study, is generated by the Mathematica notebooks that we supply as part of the SI.

\section*{Code Availability}
All the code is in the Mathematica notebooks that we supply as part of the SI.

\bibliography{references}

\end{document}